\documentclass[twocolumn, 11pt]{article}

\usepackage[utf8]{inputenc}
\usepackage[T1]{fontenc}
\usepackage{amsmath,amssymb}
\usepackage{graphicx}
\usepackage{booktabs}
\usepackage{tabularx}
\usepackage[colorlinks=true,allcolors=black]{hyperref}
\usepackage{geometry}
\usepackage{fancyhdr}
\usepackage{setspace}
\usepackage{microtype}
\usepackage{float}
\usepackage{caption}
\usepackage{url}
\usepackage[round]{natbib} 
\usepackage{titlesec}
\usepackage{enumitem}
\usepackage{stfloats}
\hypersetup{colorlinks=true, allcolors=black}

\titlespacing*{\section}{0pt}{1.5ex plus 1ex minus .2ex}{0.5ex plus .2ex}
\titlespacing*{\subsection}{0pt}{1.5ex plus 1ex minus .2ex}{0.5ex plus .2ex}
\titlespacing*{\subsubsection}{0pt}{1.5ex plus 1ex minus .2ex}{0.5ex plus .2ex}

\setlist{nosep}

\title{\textbf{The Wisdom of Deliberating AI Crowds: Does Deliberation Improve LLM-Based Forecasting?}}

\author{Paul Schneider\thanks{PRIORB, Bochum, Germany. Contact: \texttt{paul@priorb.com}} \and Amalie Schramm\footnotemark[1]}

\date{}

\begin{document}

\twocolumn[
  \begin{@twocolumnfalse}
    \maketitle
    \begin{abstract}
      Structured deliberation has been found to improve the performance of human forecasters. This study investigates whether a similar intervention—allowing LLMs to review each other's forecasts before updating—can improve accuracy in large language models (GPT-5, Claude Sonnet 4.5, Gemini Pro 2.5). Using 202 resolved binary questions from the Metaculus Q2 2025 AI Forecasting Tournament, accuracy was assessed across four scenarios: (1) diverse models with distributed information, (2) diverse models with shared information, (3) homogeneous models with distributed information, and (4) homogeneous models with shared information. Results show that the intervention significantly improves accuracy in scenario (2), reducing Log Loss by 0.020 or about 4\% in relative terms ($p = 0.017$). However, when homogeneous groups (three instances of the same model) engaged in the same process, no benefit was observed. Unexpectedly, providing LLMs with additional contextual information did not improve forecast accuracy, limiting our ability to study information pooling as a mechanism. Our findings suggest that deliberation may be a viable strategy for improving LLM forecasting.
      \vspace{2em}
    \end{abstract}
  \end{@twocolumnfalse}
]
{\renewcommand{\thefootnote}{}
\footnotetext{\hspace*{-0.5cm}\noindent\textsuperscript{*}PRIORB, Bochum, Germany. Contact: \texttt{paul@priorb.com}}
\addtocounter{footnote}{-1}}

\section*{Introduction}

Expert forecasting is the systematic elicitation of probability judgments about future events. It usually involves obtaining probability estimates from multiple experts and aggregating their judgments into a single estimate~\citep{mellers2014,armstrong2001,tetlock2005}. Probabilistic forecasts can support policy decision making and risk management across many domains, including geopolitics (e.g., election outcomes), economics, and AI safety~\citep{hanea2021,surowiecki2004,tetlock2014}.

Traditionally, forecasting relied on human experts~\citep{tetlock2005,tetlock2015}. However, recent advancements in large language models (LLMs) has sparked a new research program into whether AI systems can potentially also provide accurate forecasts~\citep{zou2022,schoenegger2024,ye2024,halawi2024}. Various studies have since explored this question and found mixed results: while some authors report that LLMs are already approaching or even exceeding human-level performance~\citep{halawi2024,schoenegger2025}, a public AI forecasting tournament showed that human expert forecasters still outperform LLM-based systems by a significant margin~\citep{metaculus2025b,metaculus2025}.

In line with benchmarks in other areas, AI forecasting results suggest that general LLM capabilities might be the most important determinant of forecast performance~\citep{brown2020,wei2022,kaplan2020,metaculus2025b}. Notwithstanding, methodological choices also matter. This includes prompt engineering, fine-tuning, retrieval strategies, and aggregation methods.

One method that has not yet been systematically tested is deliberation, i.e., the process of structured discussion and sharing of information. It has been shown to improve forecast accuracy when used by teams of human experts~\citep{hemming2018,dezecache2022}. A deliberation-like protocol (``multi-agent debate'') was also found to be effective in improving LLM performance on math and logic tasks~\citep{du2024,liang2024}. In this study, we test whether this finding extends to LLM-based forecasting systems and improves their accuracy.

\subsection*{Study Objective}

We investigate whether a deliberation-like process of sharing forecast estimates and reasoning across multiple LLM instances (``deliberation'' hereafter) improves forecast accuracy, compared to simply aggregating independent forecasts.

The hypothesis was tested under conditions that varied along two dimensions: a) model diversity (homogeneous vs.\ diverse), and b) information distribution (shared vs.\ distributed). The resulting four scenarios correspond to distinct deployment scenarios of LLM forecasting systems.

\section*{Methods}

\subsection*{Overview}

We used 202 resolved binary questions from the Metaculus Q2 2025 AI tournament. Groups of three LLMs forecasted each question in two rounds. LLMs first generated independent forecasts, then were shown their peers' forecasts and reasoning before making updated forecasts (``deliberation''). We tested four scenarios crossing model diversity (diverse vs.\ homogeneous) with information distribution (distributed vs.\ shared). Accuracy was measured using Log Loss on the median group forecast. Within each scenario, we used paired $t$-tests to compare independent vs.\ deliberative forecasts.

\subsection*{Materials: Questions, Information, and LLMs}

\subsubsection*{Forecasting Questions}

We used all 202 resolved binary questions from the Metaculus Q2 2025 AI Forecasting Benchmark~\citep{metaculus2025c}. All questions had resolved by the time of analysis, allowing us to measure forecast accuracy against ground truth. Questions spanned multiple domains including geopolitics, economics, technology, and science. Each question specified a binary outcome and a resolution date.

\subsubsection*{Information Extraction}

To manipulate information availability across agents, we extracted relevant contextual information from publicly available forecast commentary from the Metaculus platform~\citep{metaculus2025c}, accessed via the official API (\url{https://www.metaculus.com/api/}). We used an LLM (Gemini Pro 2.5) to summarise the information with the aim to construct three distinct, non-overlapping units of information. These could entail factual claims, statistical evidence, or general contextual background.

An illustrative example of the three information units is provided in Appendix~A.

All models' training cut-off dates were prior to the resolution date and question publication date, ensuring no information leakage.

\subsubsection*{LLMs}

Each forecasting task was performed by a group of three LLMs. We used three frontier models: GPT-5 (OpenAI), Claude Sonnet 4.5 (Anthropic), and Gemini Pro 2.5 (Google). Groups were configured as either homogeneous (homo), with all three LLMs being of the same type, or diverse (diverse), with one instance of each model.

The instructions provided to the models were standardized to ensure consistency across groups. For the initial (``before'') forecast, we used a modified version of the Metaculus prompting template, which has been shown to perform well in prior benchmarks~\citep{metaculus2025}. For the deliberative (``after'') forecast, we presented each model with the three initial forecasts (from all group members) along with their reasoning, and instructed them to review, contrast, and synthesise the different perspectives before updating their forecast.

Both prompt templates are provided in Appendix~B.

\subsection*{Forecast Generation Procedure}

Forecasts were generated in two stages for each question-group combination:

\textbf{Stage 1 (Independent forecasts).} Each agent in a group received the question text, resolution criteria, and their assigned information package. Agents generated forecasts and reasoning independently, without access to other agents' outputs.

\textbf{Stage 2 (Deliberative forecasts).} Each agent received the three Stage 1 forecasts and rationales from all group members. Agents were instructed to review and critique the reasoning, then provide an updated forecast. The deliberation prompt encouraged agents to identify new arguments, assess their validity, and explain any forecast revisions.

For diverse scenarios, each of the 202 questions was forecasted by one group containing all three model types (GPT-5, Sonnet, Pro). For homogeneous scenarios, questions were distributed across model types using round-robin assignment: questions 1, 4, 7, \ldots were assigned to GPT-5 groups; questions 2, 5, 8, \ldots to Sonnet groups; and questions 3, 6, 9, \ldots to Pro groups. This yielded approximately 67 questions per model type in each homogeneous condition.

Within each group, the three individual forecasts were aggregated using the median probability. The median served as the group-level forecast for both independent and deliberative stages.

\subsection*{Statistical Analysis}

We implemented a $2 \times 2$ factorial design crossing two factors:

\begin{enumerate}
\item \textbf{Model Diversity:} Homogeneous (same model) vs.\ Diverse (three different models)
\item \textbf{Information:} Shared (full information) vs.\ Distributed (unique information per LLM).
\end{enumerate}

Within each scenario, we used paired $t$-tests comparing independent vs.\ deliberative forecasts. The unit of analysis was the group-level median forecast. For diverse scenarios (group composed of three different models), we collected $n = 202$ group observations (one group per question). For homogeneous scenarios, we collected $n = 606$ group observations (three groups per question, one each of $3 \times$ GPT-5, $3 \times$ Sonnet, and $3 \times$ Pro). For the model-level breakdown (Appendix~D), questions were distributed across model types using round-robin assignment, yielding approximately 67 questions per model. The outcome metric was change in Log Loss (cross-entropy), i.e., (deliberative minus independent)~\citep{good1952}. Significance was assessed using a two-tailed $\alpha = 0.05$ threshold, not adjusted for multiple comparisons.

We conducted several secondary analyses to probe the mechanisms and boundary conditions of our findings. First, we tested whether providing information (none vs.\ partial vs.\ full) improved forecast accuracy at the independent stage, using linear regression with ``no information'' as the reference category. This isolates the effect of information from deliberation. Secondly, for homogeneous scenarios, we examined whether deliberation effects differed by model type (GPT-5, Sonnet, Pro), testing each with separate paired $t$-tests. Thirdly, we examined calibration curves to assess whether deliberation affected not just accuracy but also the calibration of probability estimates. Finally, we repeated the main analysis using Brier scores, which are less sensitive to extreme probabilities (results are reported in Appendix~E).

\subsection*{Power Estimation}

Sample size ($N = 202$) was determined by the Metaculus Q2 2025 AI Forecasting Tournament rather than a-priori power analysis. Sensitivity analyses characterising the minimum detectable effect (MDE) for each experimental condition are reported in Appendix~C.

\subsection*{Code and Data Availability}

All code, data, and analysis scripts are available at \url{https://github.com/priorb-source/delib-ai-wisdom}. Statistical analyses were conducted in Python using statsmodels~\citep{seabold2010}.

\begin{table*}[b]
\centering
\small
\caption{Effect of deliberation on forecast accuracy by scenario on Log Loss}
\label{tab:main_results}
\begin{tabular}{@{}lccccc@{}}
\toprule
Scenario & Independent & Deliberative & Change & \multicolumn{1}{c}{$t$} & \multicolumn{1}{c}{$p$} \\
 & mean (SD) & mean (SD) & mean (SD) & & \\
\midrule
Diverse models, distributed information & 0.475 (0.494) & 0.453 (0.537) & $-0.022$ (0.237) & 1.34 & 0.18 \\
Diverse models, shared information & 0.501 (0.608) & 0.481 (0.618) & $-0.020$ (0.117) & 2.41 & 0.017 \\
Homogeneous models, distributed information & 0.517 (0.612) & 0.525 (0.677) & $+0.008$ (0.308) & $-0.36$ & 0.72 \\
Homogeneous models, shared information & 0.525 (0.653) & 0.545 (0.696) & $+0.020$ (0.194) & $-1.47$ & 0.14 \\
\bottomrule
\end{tabular}
\end{table*}

\section*{Results}

The dataset comprised 202 resolved binary questions from the Metaculus Q2 2025 AI tournament. Each question was forecasted by groups of 3 agents across 4 scenarios, yielding 1,616 group-level observations (202 questions $\times$ 2 diverse scenarios $+$ 606 questions $\times$ 2 homogeneous scenarios). At the agent level, this corresponded to 2,424 individual forecasts per stage (before and after deliberation).

Table~\ref{tab:main_results} shows the effect of deliberation on forecast accuracy within each scenario. We report mean Log Loss for independent and deliberative forecasts, the paired difference (deliberative minus independent), and paired $t$-test statistics.

The results show that both diverse scenarios showed improvement after deliberation. The effect was statistically significant for Diverse models, shared information ($p = 0.017$), with a mean Log Loss reduction of 0.020, corresponding to a 4\% relative improvement in accuracy. The Diverse models, distributed information scenario showed a similar magnitude of improvement ($-0.022$) but higher variance, resulting in a non-significant result ($p = 0.18$).

Neither homogeneous scenario benefited from deliberation. Both showed slight increases in Log Loss (worsening), though neither was statistically significant.

Appendix~E reports the effect of deliberation on forecast accuracy by scenario measured in Brier score, as an alternative outcome variable. This sensitivity analysis confirmed the primary findings.

A break down of results for the homogeneous scenarios by model type is provided in Appendix~D. Gemini Pro showed the largest increases in Log Loss after deliberation ($+0.047$ and $+0.052$), while Sonnet showed slight decreases in both conditions and GPT-5 showed mixed results. However, the analysis was not powered to detect model-level differences.

\subsection*{Calibration Analysis}

\begin{figure*}[b]
\centering
\includegraphics[width=0.8\linewidth]{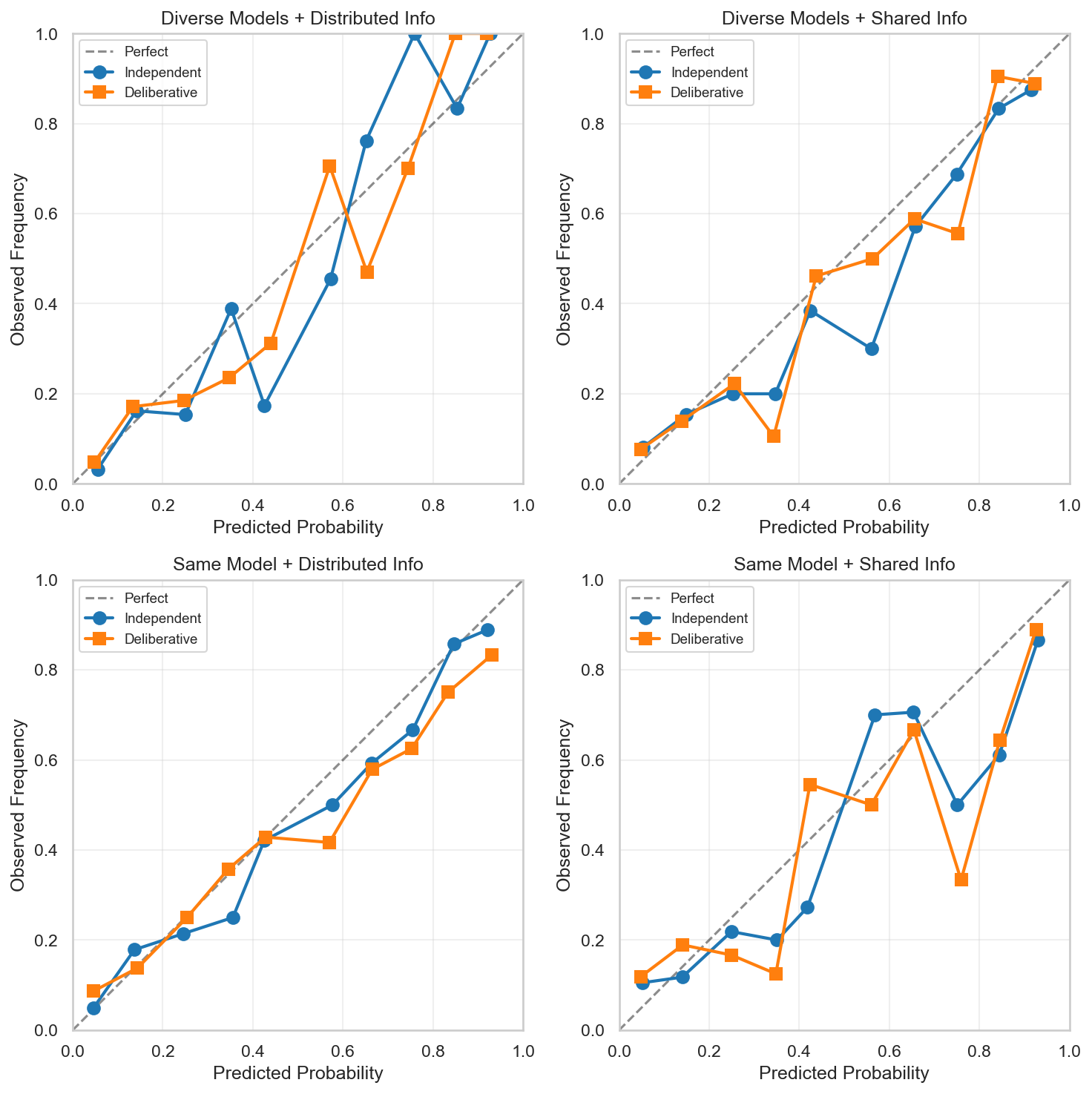}
\caption{Calibration curves for independent vs.\ deliberative forecasts, stratified by scenario}
\label{fig:calibration}
\end{figure*}

Figure~\ref{fig:calibration} shows calibration curves for all four scenarios, comparing independent (blue) and deliberative (orange) forecasts against perfect calibration (dashed diagonal). Overall, the models appear reasonably well-calibrated across conditions, with predicted probabilities generally tracking observed frequencies. Visual inspection reveals no systematic pattern distinguishing the independent from the deliberative forecasts; in most cases, the curves overlap or fluctuate without a clear directional trend.

The ``Same Model + Distributed Information'' condition appears to show the tightest alignment with the diagonal. Conversely, the ``Same Model + Shared Information'' condition exhibits the most notable deviation, largely driven by a significant outlier in the deliberative forecast at the 0.8 probability bin. Aside from this instance, however, the data do not support a strong conclusion that deliberation consistently degrades or improves calibration relative to the independent baseline.

\subsection*{Sensitivity Analysis: Does Information Affect Accuracy?}

The similar effect sizes observed for models with diverse and shared information was unexpected. We assumed that deliberation could help partly by allowing LLMs to pool information, in which case the effect of deliberation should be larger in those conditions where agents held unique facts. The absence of this pattern led us to investigate whether the information we provided was decision-relevant in the first place.

More specifically, we tested whether providing information improved forecast accuracy at the \textit{independent} stage. It was found that neither partial nor full information significantly improved forecast accuracy. Table~\ref{tab:information_effect} shows the results of linear regressions predicting Log Loss from information level, with ``no information'' as the reference category.

\begin{table}[H]
\centering
\small
\caption{Effect of information on forecast accuracy, measured as Log Loss (independent stage only)}
\label{tab:information_effect}
\begin{tabular}{@{}llccccc@{}}
\toprule
 & Predictor & $\beta$ & SE & $t$ & $p$ \\
\midrule
\multicolumn{6}{l}{\textbf{Diverse ($n = 1{,}818$)}} \\
 & Intercept (no info) & 0.567 & 0.027 & 20.66 & $<$.001 \\
 & Partial info & $-0.025$ & 0.035 & $-0.71$ & .48 \\
 & Full info & $-0.038$ & 0.039 & $-0.98$ & .33 \\
\midrule
\multicolumn{6}{l}{\textbf{Homogeneous ($n = 1{,}818$)}} \\
 & Intercept (no info) & 0.548 & 0.027 & 20.06 & $<$.001 \\
 & Partial info & $+0.001$ & 0.039 & 0.02 & .99 \\
 & Full info & $-0.008$ & 0.039 & $-0.21$ & .83 \\
\bottomrule
\end{tabular}
\end{table}

\section*{Discussion}

This study is the first to test whether structured deliberation improves forecast accuracy in LLM-based prediction systems. Our central finding is that deliberation significantly improved accuracy when diverse models collaborate: allowing models to deliberate reduced log loss by 0.020 to 0.022, which corresponds to a marginal but potentially meaningful relative improvement of around 4\% in forecast accuracy.

However, the benefit of deliberation was only found when models were diverse, i.e., when groups were composed of different models, and the improvement was only statistically significant when all models had access to the same information. When groups of LLMs were composed of three instances of the same model, deliberation yielded no accuracy gains. In fact, results even indicate a slight degradation, although not statistically significant. It could be speculated that because models share identical training data and architecture, any learned biases or reasoning errors might be correlated. Presenting such a model with ``peer'' forecasts that mirror its own might not provide any new, external perspective.

Potentially the most surprising finding is that providing additional contextual information did not improve forecast accuracy (see Table~\ref{tab:information_effect}). This result was unexpected. Metaculus tournament data suggested that the quality of retrieved information impacts forecast accuracy~\citep{metaculus2025b}. A central hypothesis of this work was indeed based on the assumption that pooling of information could be a main driver of any deliberation benefits. The most immediate explanation is that the LLM-generated summaries of Metaculus commentary may not have contained genuinely useful information. Yet, when we manually reviewed the underlying data, rationales in the scenarios in which LLMs had more information available appeared to be of higher quality than those in which less information was available. Those rationales frequently referenced external facts and data sources that, at face value, seemed decision-relevant. One could speculate that the compression of the information from comments into short information units may have led to loss of nuance or to have led models to become overconfident. Notwithstanding, these results remain counter intuitive and should be investigated further in future work.

The absence of a clear information effect has important implications for the interpreting of our findings. If additional information was generally not useful, pooling of information during deliberation cannot be expected to meaningfully improve accuracy. This likely explains why, contrary to initial expectations, the effect of deliberation under the distributed information condition was not larger than the shared information condition. In fact, both conditions showed very similar effect sizes ($\sim 0.02$ Log Loss improvement), with only the shared information condition being statistical significance, likely due to lower variance.

This study has several other limitations worth mentioning. First, we implemented a minimal deliberation protocol consisting of a single round of forecast sharing and updating. More elaborate and structured approaches might yield additional benefits. Secondly, results may be specific to the types of questions tested and the models used. Generalisation to other forecasting contexts remains to be established. Finally, the power of our study to detect true effects may have been limited by the number of questions sampled. Future work should aim to validate these findings across a larger corpus of forecasting questions.

Notwithstanding these limitations, our results both extend and qualify findings from the multi-agent debate literature in AI. \citet{du2024} and \citet{liang2024} demonstrated that debate protocols can improve LLM performance on reasoning tasks, particularly mathematics and logic problems. The present study extends this finding to the forecasting domain, where deliberation improves accuracy on real-world probabilistic forecasts. Findings reported in this study also echo patterns observed in human forecasting research. Several authors found that forecaster teams can benefit from structured interaction protocols~\citep{tetlock2015,hemming2020,dezecache2022}. \citet{hanea2021} and \citet{fraser2023} found that deliberation improved calibration even when experts had access to the same background materials. Our results suggest this pattern might transfer to groups of LLMs.

For practitioners who build forecasting systems, our findings tentatively suggest that deliberation among diverse models may improve accuracy. The marginal cost of additional API queries to implement a deliberation-like protocol is modest and may provide benefit. However, the observed improvement appears small: the 0.02 change in Log Loss corresponds to a $\sim 4\%$ relative improvement. Deliberation, if it helps, is an incremental optimisation. Base model capability likely remains the dominant factor in forecast accuracy. Benchmarking and testing system performance under real-world conditions is essential before deployment.

\section*{Conclusion}

This study found that deliberation improved accuracy in LLM-based forecasting when using diverse models: when three different frontier models reviewed each other's forecasts, accuracy improved significantly; when identical models engaged in the same process, no benefit was observed. Contrary to expectations, additional contextual information had no effect on forecast accuracy. While further research is warranted to confirm these findings and explore the role of information quality, practitioners may wish to explore deliberation-like protocols with diverse models in forecasting applications.

\section*{Funding Statement}

This study was financially supported by the Foresight Institute (\url{https://foresight.org/}).

\bibliographystyle{apalike}
\bibliography{references}

@article{good1952,
  title     = {Rational Decisions},
  author    = {Good, I. J.},
  year      = {1952},
  journal   = {Journal of the Royal Statistical Society: Series B (Methodological)},
  volume    = {14},
  number    = {1},
  pages     = {107--114},
  doi       = {10.1111/j.2517-6161.1952.tb00104.x},
  publisher = {Wiley}
}

@article{mellers2014,
  title     = {Psychological Strategies for Winning a Geopolitical Forecasting Tournament},
  author    = {Mellers, Barbara and Ungar, Lyle and Baron, Jonathan and Ramos, Jaime and Gurcay, Burcu and Fincher, Katrina and Scott, Sydney E. and Moore, Don and Atanasov, Pavel and Swift, Samuel A. and Murray, Terry and Stone, Eric and Tetlock, Philip E.},
  year      = {2014},
  journal   = {Psychological Science},
  volume    = {25},
  number    = {5},
  pages     = {1106--1115},
  doi       = {10.1177/0956797614524255},
  publisher = {SAGE Publications}
}

@book{armstrong2001,
  title     = {Principles of Forecasting: A Handbook for Researchers and Practitioners},
  editor    = {Armstrong, J. Scott},
  year      = {2001},
  publisher = {Kluwer Academic Publishers},
  address   = {Boston, MA},
  series    = {International Series in Operations Research \& Management Science},
  volume    = {30},
  doi       = {10.1007/978-0-306-47630-3},
  isbn      = {978-0-7923-7930-0}
}

@article{tetlock2014,
  title     = {Forecasting Tournaments: Tools for Increasing Transparency and Improving the Quality of Debate},
  author    = {Tetlock, Philip E. and Mellers, Barbara A. and Rohrbaugh, Nick and Chen, Eva},
  year      = {2014},
  journal   = {Current Directions in Psychological Science},
  volume    = {23},
  number    = {4},
  pages     = {290--295},
  doi       = {10.1177/0963721414534257},
  publisher = {SAGE Publications}
}

@inproceedings{zou2022,
  title     = {Forecasting Future World Events with Neural Networks},
  author    = {Zou, Andy and Xiao, Tristan and Jia, Ryan and Kwon, Joe and Mazeika, Mantas and Li, Richard and Song, Dawn and Steinhardt, Jacob and Evans, Owain and Hendrycks, Dan},
  year      = {2022},
  booktitle = {Advances in Neural Information Processing Systems},
  volume    = {35},
  note      = {Datasets and Benchmarks Track},
  url       = {https://proceedings.neurips.cc/paper_files/paper/2022/hash/aec870a6772336c15dac992c16f2e7c9-Abstract-Datasets_and_Benchmarks.html},
  eprint    = {2206.15474},
  archiveprefix = {arXiv}
}

@book{surowiecki2004,
  title     = {The Wisdom of Crowds: Why the Many Are Smarter Than the Few and How Collective Wisdom Shapes Business, Economies, Societies and Nations},
  author    = {Surowiecki, James},
  year      = {2004},
  publisher = {Doubleday},
  address   = {New York},
  isbn      = {978-0-385-50386-0}
}

@book{tetlock2015,
  title     = {Superforecasting: The Art and Science of Prediction},
  author    = {Tetlock, Philip E. and Gardner, Dan},
  year      = {2015},
  publisher = {Crown Publishers},
  address   = {New York, NY},
  isbn      = {978-0-8041-3669-3}
}

@article{kaplan2020,
  title         = {Scaling Laws for Neural Language Models},
  author        = {Kaplan, Jared and McCandlish, Sam and Henighan, Tom and Brown, Tom B. and Chess, Benjamin and Child, Rewon and Gray, Scott and Radford, Alec and Wu, Jeffrey and Amodei, Dario},
  year          = {2020},
  journal       = {arXiv preprint},
  eprint        = {2001.08361},
  archiveprefix = {arXiv},
  primaryclass  = {cs.LG},
  url           = {https://arxiv.org/abs/2001.08361}
}

@inproceedings{brown2020,
  title     = {Language Models are Few-Shot Learners},
  author    = {Brown, Tom and Mann, Benjamin and Ryder, Nick and Subbiah, Melanie and Kaplan, Jared D. and Dhariwal, Prafulla and Neelakantan, Arvind and Shyam, Pranav and Sastry, Girish and Askell, Amanda and Agarwal, Sandhini and Herbert-Voss, Ariel and Krueger, Gretchen and Henighan, Tom and Child, Rewon and Ramesh, Aditya and Ziegler, Daniel M. and Wu, Jeffrey and Winter, Clemens and Hesse, Christopher and Chen, Mark and Sigler, Eric and Litwin, Mateusz and Gray, Scott and Chess, Benjamin and Clark, Jack and Berner, Christopher and McCandlish, Sam and Radford, Alec and Sutskever, Ilya and Amodei, Dario},
  year      = {2020},
  booktitle = {Advances in Neural Information Processing Systems},
  volume    = {33},
  pages     = {1877--1901},
  url       = {https://proceedings.neurips.cc/paper/2020/hash/1457c0d6bfcb4967418bfb8ac142f64a-Abstract.html}
}

@article{wei2022,
  title         = {Emergent Abilities of Large Language Models},
  author        = {Wei, Jason and Tay, Yi and Bommasani, Rishi and Raffel, Colin and Zoph, Barret and Borgeaud, Sebastian and Yogatama, Dani and Bosma, Maarten and Zhou, Denny and Metzler, Donald and Chi, Ed H. and Hashimoto, Tatsunori and Vinyals, Oriol and Liang, Percy and Dean, Jeff and Fedus, William},
  year          = {2022},
  journal       = {Transactions on Machine Learning Research},
  eprint        = {2206.07682},
  archiveprefix = {arXiv},
  url           = {https://openreview.net/forum?id=yzkSU5zdwD}
}

@inproceedings{seabold2010,
  title     = {Statsmodels: Econometric and Statistical Modeling with Python},
  author    = {Seabold, Skipper and Perktold, Josef},
  year      = {2010},
  booktitle = {Proceedings of the 9th Python in Science Conference},
  pages     = {92--96},
  doi       = {10.25080/Majora-92bf1922-011},
  publisher = {SciPy}
}

@inproceedings{du2024,
  title         = {Improving Factuality and Reasoning in Language Models through Multiagent Debate},
  author        = {Du, Yilun and Li, Shuang and Torralba, Antonio and Tenenbaum, Joshua B. and Mordatch, Igor},
  year          = {2024},
  booktitle     = {Proceedings of the 41st International Conference on Machine Learning},
  eprint        = {2305.14325},
  archiveprefix = {arXiv},
  url           = {https://arxiv.org/abs/2305.14325}
}

@inproceedings{liang2024,
  title     = {Encouraging Divergent Thinking in Large Language Models through Multi-Agent Debate},
  author    = {Liang, Tian and He, Zhiwei and Jiao, Wenxiang and Wang, Xing and Wang, Yan and Wang, Rui and Yang, Yujiu and Shi, Shuming and Tu, Zhaopeng},
  year      = {2024},
  booktitle = {Proceedings of the 2024 Conference on Empirical Methods in Natural Language Processing},
  pages     = {17889--17904},
  doi       = {10.18653/v1/2024.emnlp-main.992},
  publisher = {Association for Computational Linguistics}
}

@book{tetlock2005,
  title     = {Expert Political Judgment: How Good Is It? How Can We Know?},
  author    = {Tetlock, Philip E.},
  year      = {2005},
  publisher = {Princeton University Press},
  address   = {Princeton, NJ},
  isbn      = {978-0-691-12302-8}
}

@article{hemming2020,
  title     = {Improving Expert Forecasts in Reliability: Application and Evidence for Structured Elicitation Protocols},
  author    = {Hemming, Victoria and Armstrong, Nicholas and Burgman, Mark A. and Hanea, Anca M.},
  year      = {2020},
  journal   = {Quality and Reliability Engineering International},
  volume    = {36},
  number    = {2},
  pages     = {623--641},
  doi       = {10.1002/qre.2596},
  publisher = {Wiley}
}

@article{dezecache2022,
  title     = {Democratic Forecast: Small Groups Predict the Future Better Than Individuals and Crowds},
  author    = {Dezecache, Guillaume and Dockendorff, Martin and Ferreiro, Dardo N. and Deroy, Ophelia and Bahrami, Bahador},
  year      = {2022},
  journal   = {Journal of Experimental Psychology: Applied},
  volume    = {28},
  number    = {3},
  pages     = {525--537},
  doi       = {10.1037/xap0000424},
  publisher = {American Psychological Association}
}

@article{fraser2023,
  title     = {Predicting Reliability through Structured Expert Elicitation with the {repliCATS} (Collaborative Assessments for Trustworthy Science) Process},
  author    = {Fraser, Hannah and Bush, Martin and Wintle, Bonnie C. and Mody, Fallon and Smith, Eden T. and Hanea, Anca M. and Gould, Elliot and Hemming, Victoria and Hamilton, Daniel G. and Rumpff, Libby and Wilkinson, David P. and Pearson, Ross and Singleton Thorn, Felix and Gray, Charles T. and Head, Andrew and Ross, Melissa and Groenewegen, Rebecca and Marcoci, Alexandru and Vercammen, Ans and Parker, Timothy H. and Hoekstra, Rink and Nakagawa, Shinichi and Mandel, David R. and van Ravenzwaaij, Don and McBride, Marissa and Sinnott, Richard O. and Vesk, Peter and Burgman, Mark and Fidler, Fiona},
  year      = {2023},
  journal   = {PLOS ONE},
  volume    = {18},
  number    = {1},
  pages     = {e0274429},
  doi       = {10.1371/journal.pone.0274429},
  publisher = {Public Library of Science}
}

@article{hemming2018,
  title     = {Eliciting Improved Quantitative Judgements Using the {IDEA} Protocol: A Case Study in Natural Resource Management},
  author    = {Hemming, Victoria and Walshe, Terry V. and Hanea, Anca M. and Fidler, Fiona and Burgman, Mark A.},
  year      = {2018},
  journal   = {PLOS ONE},
  volume    = {13},
  number    = {6},
  pages     = {e0198468},
  doi       = {10.1371/journal.pone.0198468},
  publisher = {Public Library of Science}
}

@online{metaculus2025,
  title   = {{Q1} {AI} Benchmarking Results: Pro Forecasters Crush Bots},
  author  = {{Metaculus}},
  year    = {2025},
  url     = {https://www.metaculus.com/notebooks/38673/q1-ai-benchmarking-results/},
  urldate = {2025-12-26}
}

@online{metaculus2025b,
  title   = {Fall 2025 {AI} Forecasting Benchmark Tournament},
  author  = {{Metaculus}},
  year    = {2025},
  url     = {https://www.metaculus.com/tournament/fall-aib-2025/},
  urldate = {2025-12-26}
}

@online{metaculus2025c,
  title   = {{AI} Forecasting Benchmark Tournament -- 2025 {Q2}},
  author  = {{Metaculus}},
  year    = {2025},
  url     = {https://www.metaculus.com/tournament/aibq2/},
  urldate = {2025-12-26}
}

@article{ye2024,
  title         = {{Mirai}: Evaluating {LLM} Agents for Event Forecasting},
  author        = {Ye, Chenchen and Hu, Ziniu and Deng, Yihe and Huang, Zijie and Ma, Mingyu Derek and Zhu, Yanqiao and Wang, Wei},
  year          = {2024},
  journal       = {arXiv preprint},
  eprint        = {2407.01231},
  archiveprefix = {arXiv},
  primaryclass  = {cs.CL},
  url           = {https://arxiv.org/abs/2407.01231}
}

@article{schoenegger2024,
  title     = {Wisdom of the Silicon Crowd: {LLM} Ensemble Prediction Capabilities Rival Human Crowd Accuracy},
  author    = {Schoenegger, Philipp and Tuminauskaite, Indre and Park, Peter S. and Valdece Sousa Bastos, Rafael and Tetlock, Philip E.},
  year      = {2024},
  journal   = {Science Advances},
  volume    = {10},
  number    = {45},
  pages     = {eadp1528},
  doi       = {10.1126/sciadv.adp1528},
  publisher = {American Association for the Advancement of Science}
}

@article{hanea2021,
  title     = {Mathematically Aggregating Experts' Predictions of Possible Futures},
  author    = {Hanea, Anca M. and Wilkinson, David P. and McBride, Marissa and Lyon, Aidan and van Ravenzwaaij, Don and Singleton Thorn, Felix and Gray, Charles and Mandel, David R. and Willcox, Aaron and Gould, Elliot and Smith, Eden T. and Mody, Fallon and Bush, Martin and Fidler, Fiona and Fraser, Hannah and Wintle, Bonnie C.},
  year      = {2021},
  journal   = {PLOS ONE},
  volume    = {16},
  number    = {9},
  pages     = {e0256919},
  doi       = {10.1371/journal.pone.0256919},
  publisher = {Public Library of Science}
}

@inproceedings{halawi2024,
  title     = {Approaching Human-Level Forecasting with Language Models},
  author    = {Halawi, Danny and Zhang, Fred and Chen, Yueh-Han and Steinhardt, Jacob},
  year      = {2024},
  booktitle = {Advances in Neural Information Processing Systems},
  volume    = {37},
  url       = {https://proceedings.neurips.cc/paper_files/paper/2024/hash/5a5acfd0876c940d81619c1dc60e7748-Abstract-Conference.html},
  eprint    = {2402.18563},
  archiveprefix = {arXiv}
}

@article{schoenegger2025,
  title     = {{AI}-Augmented Predictions: {LLM} Assistants Improve Human Forecasting Accuracy},
  author    = {Schoenegger, Philipp and Park, Peter S. and Karger, Ezra and Trott, Sean and Tetlock, Philip E.},
  year      = {2025},
  journal   = {ACM Transactions on Interactive Intelligent Systems},
  volume    = {15},
  number    = {1},
  pages     = {1--25},
  doi       = {10.1145/3707649},
  publisher = {Association for Computing Machinery}
}

\clearpage
\onecolumn

\appendix
\part*{Supplementary Materials}
\setcounter{section}{0}
\setcounter{figure}{0}
\setcounter{table}{2}
\renewcommand{\thesection}{S\arabic{section}}
\renewcommand{\thefigure}{S\arabic{figure}}
\renewcommand{\thetable}{S\arabic{table}}

\section{Information Unit Examples}

\begin{verbatim}
questionTitle: "Will initial jobless claims for the week ended 
June 21, 2025 exceed 220,000?"

questionDescription: "According to the resolution source, 'An 
initial claim is a claim filed by an unemployed individual after 
a separation from an employer. The claim requests a determination 
of basic eligibility for the Unemployment Insurance program.'"

questionResolutionCriteria: "This question resolves as Yes if 
initial jobless claims for the week ended June 21, 2025 is 
greater than 220,000 according to FRED"

information_1: "According to the U.S. Department of Labor, 
initial jobless claims for the week ending May 31, 2025, rose 
by 8,000 to 247,000, the highest level in eight months. The 
figure for the prior week was revised to 239,000. The four-week 
moving average increased to 235,000, the highest since November 
2021. For the week ending May 24, the number of continuing 
claims was 1.904 million, a slight decrease of 3,000 from the 
previous week."

information_2: "Broader labor market indicators suggest a 
softening trend. A report from Challenger, Gray & Christmas 
showed that U.S.-based employers announced 93,816 job cuts in 
May, which is 47% higher than in May 2024. Separately, the ADP 
employment report revealed that only 37,000 jobs were created 
in the private sector in May, the lowest figure in a year. Both 
the Federal Reserve's Beige Book report and an Institute for 
Supply Management (ISM) survey pointed to weakening conditions, 
with widespread comments about economic uncertainty delaying 
new hiring."

information_3: "The recent rise in jobless claims was not 
uniform across the country. The largest increases in initial 
claims for the week ending May 31 were in Michigan, Nebraska, 
California, Florida, and Virginia. The largest decreases were 
reported in Massachusetts, Illinois, Texas, Washington, and New 
York. Several major corporations have announced layoffs in 2025, 
including Procter & Gamble, Microsoft, Citigroup, Walmart, Meta, 
and Disney."
\end{verbatim}

\section{Prompts}

\textbf{Modified Metaculus template for Round 1 (independent):}

\begin{verbatim}
You are a professional forecaster interviewing for a job.
Your interview question is: {{questionTitle}}
Question background: {{questionDescription}}

This question's outcome will be determined by the specific 
criteria below. These criteria have not yet been satisfied: 
{{questionResolutionCriteria}}

{{question.questionFinePrint}}

Your research assistant's report says: {{information}}

Today is {{question.date}}.

Before answering you think: (a) The time left until the 
outcome to the question is known. (b) The status quo outcome 
if nothing changed. (c) A brief description of a scenario that 
results in a No outcome. (d) A brief description of a scenario 
that results in a Yes outcome. (e) You write your rationale 
remembering that good forecasters put extra weight on the 
status quo outcome since the world changes slowly most of the 
time. Explain your reasoning and the evidence behind your 
forecast in detail. Summarise information your received from 
your research assistant that influences your forecast (if any). 
Explain why your forecast is not higher, and why it is not 
lower. Outline what would need to be true for you to update 
your forecast in either direction. (f) The last thing you write 
is your final probabilistic forecast as a number between 0 
and 100.

OUTPUT SCHEMA { 
  "time_left_until_outcome_known": "string", 
  "status_quo_outcome": "string", 
  "no_outcome_scenario": "string", 
  "yes_outcome_scenario": "string", 
  "rationale": "string", 
  "probability": "number" (0-100) 
}
\end{verbatim}

\textbf{Deliberation prompt for Round 2:}

\begin{verbatim}
You are now in a deliberation phase with two other expert 
forecasters. Please review their analyses:

Forecaster 2's Analysis
{{forecaster2_rationale}} Forecast: {{forecaster2_probability}}%

Forecaster 3's Analysis
{{forecaster3_rationale}} Forecast: {{forecaster3_probability}}%

Consider their reasoning and any new information or arguments 
carefully:

- What evidence or arguments did they raise that you hadn't 
  considered?
- Do you find their reasoning convincing? Why or why not?
- Should you update your forecast based on their input? If so, 
  how much? If not, why not?

Weigh your previous analysis and critically review your own 
reasoning and evidence in light of any new information or 
arguments, as if you were participating in a structured 
deliberation process.

Based on your thoughtful analysis, provide a clear and concise 
review of all the arguments and information you have considered, 
your updated rationale, and your updated forecast. Do not feel 
obligated to update your forecast if you do not think it is 
warranted.

Provide your updated analysis and forecast.

OUTPUT SCHEMA { 
  "review": "string (your thoughts on the other forecasters' 
             reasoning)", 
  "rationale": "string (your updated reasoning; if you change 
               your forecast, explain why and how much; if not, 
               explain why not)", 
  "probability": "number" (0-100) 
}
\end{verbatim}

\section{Minimum Detectable Effects}

The sample size for this study ($N = 202$ questions) was determined externally by the Metaculus Q2 2025 AI Forecasting Tournament rather than by a-priori power analysis. We therefore conducted sensitivity analyses to characterise the minimum detectable effect (MDE) for each scenario, given the fixed $N$ and observed variance.

For a paired $t$-test with $N = 202$ observations at $\alpha = 0.05$ (two-sided), achieving 80\% power requires a Cohen's $d$ of approximately 0.198. The MDE in raw units (Log Loss) is then calculated as: MDE = $d \times$ SD, where SD is the observed standard deviation of paired differences for each scenario.

\begin{table}[H]
\centering
\small
\caption{Minimum Detectable Effects by Scenario}
\label{tab:mde}
\begin{tabular}{@{}lcccc@{}}
\toprule
Scenario & SD of Change & MDE (80\% power) & Observed Effect & $p$-value \\
\midrule
Diverse models, shared information & 0.117 & 0.023 & $-0.020$ & 0.017 \\
Diverse models, distributed information & 0.237 & 0.047 & $-0.022$ & 0.182 \\
Same model, shared information & 0.194 & 0.038 & $+0.020$ & 0.144 \\
Same model, distributed information & 0.308 & 0.061 & $+0.008$ & 0.717 \\
\bottomrule
\end{tabular}
\end{table}

The distributed information condition introduces additional variance into the deliberation process. When agents hold different information, their independent forecasts vary more, and consequently their post-deliberation updates also vary more. This variance increase is itself substantively meaningful: it suggests that information asymmetry creates noise that may partially obscure any deliberation benefit. Studies examining deliberation with distributed information should anticipate approximately $2 \times$ higher variance than shared-information conditions. To achieve equivalent power, such studies would require either larger samples or larger true effects. At $N = 202$, the Diverse models, distributed information condition was adequately powered only to detect effects of $\sim 0.05$ Log Loss or larger—roughly 2.5$\times$ the effect we observed.

\begin{figure}[H]
\centering
\includegraphics[width=0.95\linewidth]{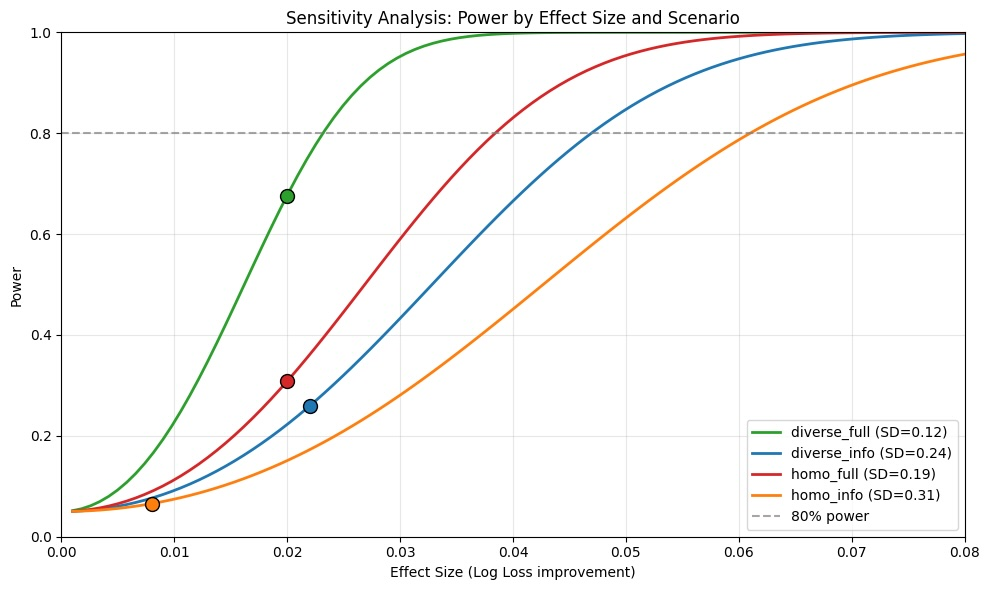}

\vspace{0cm}
\begin{center}
  \textbf{Figure S3:} Sensitivity curves showing statistical power as a function of effect size for each scenario. Dots indicate observed effects; horizontal line marks 80\% power threshold.
\end{center}
\end{figure}

\section{Model-Level Breakdown (Homogeneous Scenarios)}

\begin{table}[H]
\centering
\small
\caption{Deliberation effects by model type (homogeneous scenarios only)}
\label{tab:model_breakdown}
\begin{tabular}{@{}lcccccccc@{}}
\toprule
Scenario & Model & $n$ & Independent & Deliberative & Change & $t$ & $p$ \\
\midrule
Same model, distributed info. & GPT-5 & 67 & 0.483 & 0.476 & $-0.007$ & 0.56 & 0.58 \\
Same model, distributed info. & Sonnet & 67 & 0.440 & 0.422 & $-0.017$ & 0.65 & 0.52 \\
Same model, distributed info. & Pro & 68 & 0.627 & 0.675 & $+0.047$ & $-0.83$ & 0.41 \\
Same model, shared info. & GPT-5 & 67 & 0.485 & 0.494 & $+0.009$ & $-0.99$ & 0.33 \\
Same model, shared info. & Sonnet & 67 & 0.437 & 0.435 & $-0.002$ & 0.27 & 0.79 \\
Same model, shared info. & Pro & 68 & 0.651 & 0.703 & $+0.052$ & $-1.35$ & 0.18 \\
\bottomrule
\end{tabular}

\vspace{1ex}
\small \textit{Note:} Questions were distributed across model types using round-robin assignment. Values show mean Log Loss.
\end{table}

\section{Brier Score Sensitivity Analysis}

Table~\ref{tab:brier} below shows the primary analysis repeated using Brier scores (instead of Log Loss). Brier scores are less sensitive to extreme probability estimates and may provide a complementary measure of forecast accuracy.

\begin{table}[H]
\centering
\small
\caption{Effect of deliberation on forecast accuracy by scenario (Brier Score)}
\label{tab:brier}
\begin{tabular}{@{}lccccc@{}}
\toprule
Scenario & Independent & Deliberative & Change & $t$ & $p$ \\
 & mean (SD) & mean (SD) & mean (SD) & & \\
\midrule
Diverse models, distributed information & 0.153 (0.188) & 0.145 (0.201) & $-0.008$ (0.102) & 1.14 & 0.26 \\
Diverse models, shared information & 0.162 (0.221) & 0.153 (0.220) & $-0.009$ (0.051) & 2.47 & 0.014 \\
Homogeneous models, distributed information & 0.169 (0.214) & 0.170 (0.234) & $+0.001$ (0.123) & $-0.12$ & 0.90 \\
Homogeneous models, shared information & 0.171 (0.240) & 0.177 (0.250) & $+0.007$ (0.061) & $-1.56$ & 0.12 \\
\bottomrule
\end{tabular}

\vspace{1ex}
\small \textit{Note:} $n = 202$ for all scenarios. Change = Deliberative minus Independent; negative values indicate improvement.
\end{table}

\end{document}